\documentclass[sigplan,nonacm]{acmart}
\usepackage{epsfig,endnotes}
\usepackage[T1]{fontenc}
\usepackage[scaled=0.75]{beramono}
\usepackage{amsmath,amsthm}
\usepackage{newtxtext,newtxmath}  %
\usepackage{simplebnf}
\usepackage[ruled,vlined,noend]{algorithm2e}
\usepackage{listings}
\usepackage{xcolor}
\usepackage{colortbl}
\usepackage{graphicx}
\usepackage{subcaption}
\usepackage{wrapfig}
\usepackage[font={small},labelfont=bf]{caption}
\usepackage{booktabs}
\usepackage{adjustbox}
\usepackage{datetime}
\usepackage{enumitem}
\usepackage{multirow}
\usepackage{minted}
\usepackage[binary-units=true]{siunitx}
\usepackage{pifont}
\usepackage{tikz}
\usetikzlibrary{tikzmark,calc}
\usepackage{cleveref}
\usepackage{algorithmicx}
\usepackage[normalem]{ulem}
\usepackage[colorinlistoftodos]{todonotes}
\usepackage{listings}
\usepackage{subcaption}
\usepackage[dvipsnames]{xcolor}
\definecolor{codegreen}{rgb}{0,0.6,0}
\definecolor{codegray}{rgb}{0.5,0.5,0.5}
\definecolor{codepurple}{rgb}{0.58,0,0.82}
\definecolor{backcolour}{rgb}{0.95,0.95,0.92}

\lstdefinestyle{mystyle}{
    backgroundcolor=\color{backcolour},   
    commentstyle=\color{codegreen},
    keywordstyle=\color{NavyBlue},
    numberstyle=\tiny\color{codegray},
    stringstyle=\color{codepurple},
    basicstyle=\ttfamily\footnotesize,
    breakatwhitespace=false,         
    breaklines=true,                 
    captionpos=b,                    
    keepspaces=true,                 
    numbers=left,                    
    numbersep=5pt,                  
    showspaces=false,                
    showstringspaces=false,
    showtabs=false,                  
    tabsize=2
}

\usepackage{tabularx}

\graphicspath{{./}{figure/}}

\setminted{style=xcode,breaklines=true,breaksymbolleft=,fontsize=\footnotesize}
\DeclareMathAlphabet{\mathcal}{OMS}{cmsy}{m}{n}

\newcommand{\sys}{\mbox{\textsc{Scalify}}\xspace} %

\definecolor{Brown}{RGB}{124, 66, 0}
\newcounter{findings}

\DeclareSIUnit{\operation}{op}
\newcolumntype{R}[2]{%
  >{\adjustbox{angle=#1,lap=\width-(#2)}\bgroup}%
  l%
  <{\egroup}%
}

\newcommand{\locpinpoint}{\ding{248}}
\newcommand{\locfunc}{\ding{91}}

\crefname{section}{\S}{\S\S}
\crefname{subsection}{\S}{\S\S}
\crefformat{section}{\S#2#1#3}
\crefformat{subsection}{\S#2#1#3}

\lstdefinelanguage{json}{
    basicstyle=\normalfont\ttfamily,
    numbers=left,
    numberstyle=\scriptsize,
    stepnumber=1,
    numbersep=8pt,
    showstringspaces=false,
    breaklines=true,
    frame=lines,
    backgroundcolor=\color{background},
    literate=
     *{0}{{{\color{numb}0}}}{1}
      {1}{{{\color{numb}1}}}{1}
      {2}{{{\color{numb}2}}}{1}
      {3}{{{\color{numb}3}}}{1}
      {4}{{{\color{numb}4}}}{1}
      {5}{{{\color{numb}5}}}{1}
      {6}{{{\color{numb}6}}}{1}
      {7}{{{\color{numb}7}}}{1}
      {8}{{{\color{numb}8}}}{1}
      {9}{{{\color{numb}9}}}{1}
      {:}{{{\color{punct}{:}}}}{1}
      {,}{{{\color{punct}{,}}}}{1}
      {\{}{{{\color{delim}{\{}}}}{1}
      {\}}{{{\color{delim}{\}}}}}{1}
      {[}{{{\color{delim}{[}}}}{1}
      {]}{{{\color{delim}{]}}}}{1},
}

\begin{document}
\hypersetup{
  colorlinks=true,
  linkcolor=purple,
  citecolor=blue,
  urlcolor=magenta,
}

\pagestyle{plain}

\title{ 
Verifying Computational Graphs in Production-Grade Distributed Machine Learning Frameworks
}

\author{Kahfi S. Zulkifli}
\affiliation{%
  \institution{University of Virginia}
  \country{USA}
  }

  \author{Wenbo Qian}
\affiliation{%
  \institution{Northeastern University}
  \country{USA}
  }

  \author{Shaowei Zhu}
\affiliation{%
  \institution{Amazon Web Services}
  \country{USA}
  }

    \author{Yuan Zhou}
\affiliation{%
  \institution{Amazon Web Services}
  \country{USA}
  }

    \author{Zhen Zhang}
\affiliation{%
  \institution{Amazon Web Services}
  \country{USA}
  }

  \author{Chang Lou}
\affiliation{%
  \institution{University of Virginia}
  \country{USA}
  }

\renewcommand{\shortauthors}{Zulkifli et al.}

\begin{abstract}

Modern machine learning frameworks support very large models by incorporating 
parallelism and optimization techniques.
Yet, these very techniques add new layers of complexity, introducing 
silent errors that severely degrade model performance.  
Existing solutions are either ad hoc or too costly for production.

We present \sys, a lightweight framework that exposes silent errors by verifying semantic equivalence of computational graphs using equality saturation and Datalog-style reasoning.
\sys partitions graphs with parallel rewriting and layer memoization, reuses rewrite templates, and augments equality saturation with relational reasoning and symbolic bijection inference. It further localizes discrepancies to precise code sites, turning verification results into actionable debugging guidance.
\sys verifies models as large as
Llama-3.1-405B within minutes on a commodity machine and
exposed five unknown bugs in Amazon production machine learning frameworks.

\end{abstract}

\maketitle 
\pagenumbering{arabic}

\section{Introduction}
\label{sec:intro}

Machine learning research today has increasingly become a race to scale, 
where larger models consistently outperform smaller counterparts across diverse tasks.
From language models like GPT~\cite{gpt32020nips} to vision transformers~\cite{khan2022transformers}, 
performance gains are often linked to model size, with billions of parameters enabling more 
expressive representations and generalization.
The largest Llama-3 model~\cite{llama3.1} boasts 405 billion parameters, 
and DeepSeekV3~\cite{deepseekv3} exceeds 671 billion. 

For years, scaling techniques, such as distributed training, memory optimizations, 
and hardware accelerations, have been the primary focus to 
support ever-larger models. However, as models reach unprecedented scales, a new reliability crisis is 
emerging. Developers increasingly observe silent errors~\cite{tambon2023silentbugsdeeplearning, 
optimizebugstudy2023icse, compilerstudy2021fse,nnsmith2023asplos}—severe performance 
degradation in trained models without triggering explicit error signals.
For example, they are often introduced by bugs~\cite{gradientbug1,gradientbug2,gradientbug3} 
in distributed training when synchronizing between devices.
Thus, scaling alone is not enough; ensuring robustness becomes the next critical challenge.

Yet, detecting and diagnosing silent errors in ML models remains a crucial but elusive challenge.
Deep learning models emerge from a complex pipeline involving ML 
frameworks~\cite{pytorch,tensorflow}, graph optimizers~\cite{xla,tvm2018osdi}, 
schedulers~\cite{bagpipe2023sosp,mast2024osdi}, and backends~\cite{cuda}. Bugs can 
lurk at any of these layers, making troubleshooting a daunting task. Worse yet, silent 
errors do not just stem from software bugs—they can arise from numerical 
instabilities~\cite{numericalerror2020pldi} or even faulty hardware~\cite{drdna2024asplos}.

Despite extensive efforts in testing of ML frameworks~\cite{cradle2019icse,diffchase2019IJCAI} and 
compilers~\cite{nnsmith2023asplos,tzer2022oopsla,polyjuice2024oopsla},
they are far from sufficient. Bugs often surface only after severe performance 
degradation occurs, making pre-training checking an urgent necessity. Yet, developers today 
still rely on an \emph{ad hoc} approach—manually extracting and comparing intermediate activation or gradient 
tensor values at different locations. This process is not only tedious but also unreliable and fragile, as 
floating-point computations naturally introduce small discrepancies whose magnitude depend on multiple factors
including hardware platform, compute kernels used, tensor shapes, and the smoothness of the computation itself, which makes numerical comparisons 
challenging. Even when discrepancies are detected, pinpointing 
the root cause remains a painstaking effort, leaving developers struggling with uncertainty. A more systematic and robust detection solution is urgently needed.

In this work, we argue that verifying training models at the \emph{semantic} level 
rather than relying on arithmetic approximations—is not only feasible but essential.
Modern machine learning methods can be expressed in computational graphs,
which capture the structured flow of data and transformations throughout training. 
By analyzing the semantic equivalence between the original and transformed graphs,
we can move beyond imprecise numerical comparisons and systematically detect and diagnose silent errors.

\emph{How can we compare two different ML computational graphs?}
Recent work~\cite{eleftheriadis2022neuralnetworkequivalencechecking, trainverifySOSP2025} has explored SMT-based verification methods. 
For example, TrainVerify~\cite{trainverifySOSP2025} symbolizes computational graphs by converting tensors/operations into logical formulas and using an SMT solver to 
verify the equivalence of such formulas. 
However, this approach faces practicality challenges in production. The reasoning for nonlinear real arithmetic is expensive; in practice the approach relies on solver tactics that can be fragile, require workload-specific reductions (\emph{e.g.,} ``shape reduction'' that assumes linearity) to be tractable, and can still take days on very large models. The approach also depends on maintaining fine-grained tensor lineage between logical and parallel graphs---information many production frameworks do not preserve---and, when a proof fails, the resulting unsat core can be hard to map back to the program site that breaks equivalence.

In this work, we ask a central question:
\emph{Is there a more lightweight, flexible, and actionable approach to verify ML computational graphs?}
While SMT-based methods are theoretically powerful, we argue they are unnecessarily heavy for this task since large portions of the graphs are structurally identical across transformations.
In practice, developers often rely on a different process for checking the correctness of computational graphs: manually rewriting and comparing two graphs step by step according to computation semantics until they either converge or remain divergent---a tedious and error-prone effort even on small graphs.

Inspired by this observation, we propose a novel approach to verify graph-equivalence via e-graphs~\cite{es2009popl, egg2021popl}: generic rewrite rules capture families of tensor sizes and parallelization configurations, equality saturation searches equivalence spaces directly on the IR graph, and discrepancies naturally pinpoint offending subgraphs/operations for bug localization. 

It is not straightforward to directly apply this approach to ML computational graphs.
First, modern ML models are large and extremely complex. Naively constructing e-graphs by inserting all nodes and including all possible rewrites  easily leads to exponential blow up in time and memory usage. Second, we must balance rule generality and practicality: highly general rules capture more bugs but cause significant overhead, while overly specific rules bloat the system with limited reuse. Third, modern pipelines introduce layout heterogeneity, where many reshape/transpose sequences share shapes but differ semantically, requiring careful reasoning to avoid false matches. Finally, a simple equivalence verdict is not useful---developers still need bug location in codes to act on verification results.

Thus, we introduce \sys, a lightweight framework to verify computational graph equivalence in production distributed ML frameworks.  
Our design tackles these challenges with a set of new techniques.
\sys control graph-scale explosion through partitioning, parallel rewriting, and layer memoization, enabling scalability to deep models. To balance rule generality and practicality, we introduce reusable rewrite templates that capture common patterns. For layout heterogeneity, we augment e-graph rewriting with relational reasoning and symbolic bijection inference, ensuring precise semantic alignment of tensor transformations. Finally, we make verification results actionable by localizing discrepancies back to the exact operations and code locations, providing developers with clear guidance to resolve bugs.

We evaluate \sys on large real-world models and production frameworks. 
Our tool verifies models as large as Llama-3.1-405B within minutes on a commodity machine, showing scalability across multiple parallelism techniques including tensor, sequence, and expert parallelism. \sys uncovered five previously unknown bugs in Amazon’s Transformers NeuronX~\cite{transformersneuronx} and NeuronX Distributed~\cite{nxd} frameworks, in addition to detecting and localizing 17 out of 19 reproduced bugs from prior studies.

In summary, this paper makes the following contributions:

 \begin{itemize}[noitemsep, topsep=0pt, partopsep=0pt, leftmargin=*]
\item We propose a novel approach for verifying the semantic equivalence of ML models using equality saturation.
  \item We develop \sys, a verification tool that integrates a set of  techniques to apply the proposed approach to large-scale ML models in production.
  \item We evaluate \sys on two production ML frameworks from Amazon, demonstrating its effectiveness in verifying large real-world models and uncovering previously unknown bugs.

 \end{itemize}

\section{Background}
\label{sec:motivation}

\subsection{Computational Graph}
Modern ML frameworks apply a series of transformations to convert high-level tensor programs into low-level codes.
Common transformations include parallelization techniques~\cite{gpipe2019nips,shoeybi2020megatronlmtrainingmultibillionparameter} that distribute workloads across multiple accelerators (\emph{e.g.,} data, model, and pipeline parallelism);
operator fusion~\cite{hidet2023asplos,relaxed2019mlsys,deepcuts2021pldi}, which merges compatible operations (\emph{e.g.,} convolution plus bias and activation) into a single kernel to reduce overhead;  
and algebraic rewrites~\cite{taso2019sosp,pet2021osdi,unity2022osdi,superoptimization2021mlsys} like constant folding and redundant-operator elimination.  
These transformations are generally complex and easily include  bugs, introducing silent errors in the compiled programs. %

\setlength{\intextsep}{2pt}%
\setlength{\columnsep}{5pt}%
\begin{figure}[t]
    \centering
      \includegraphics[width=0.5\textwidth]{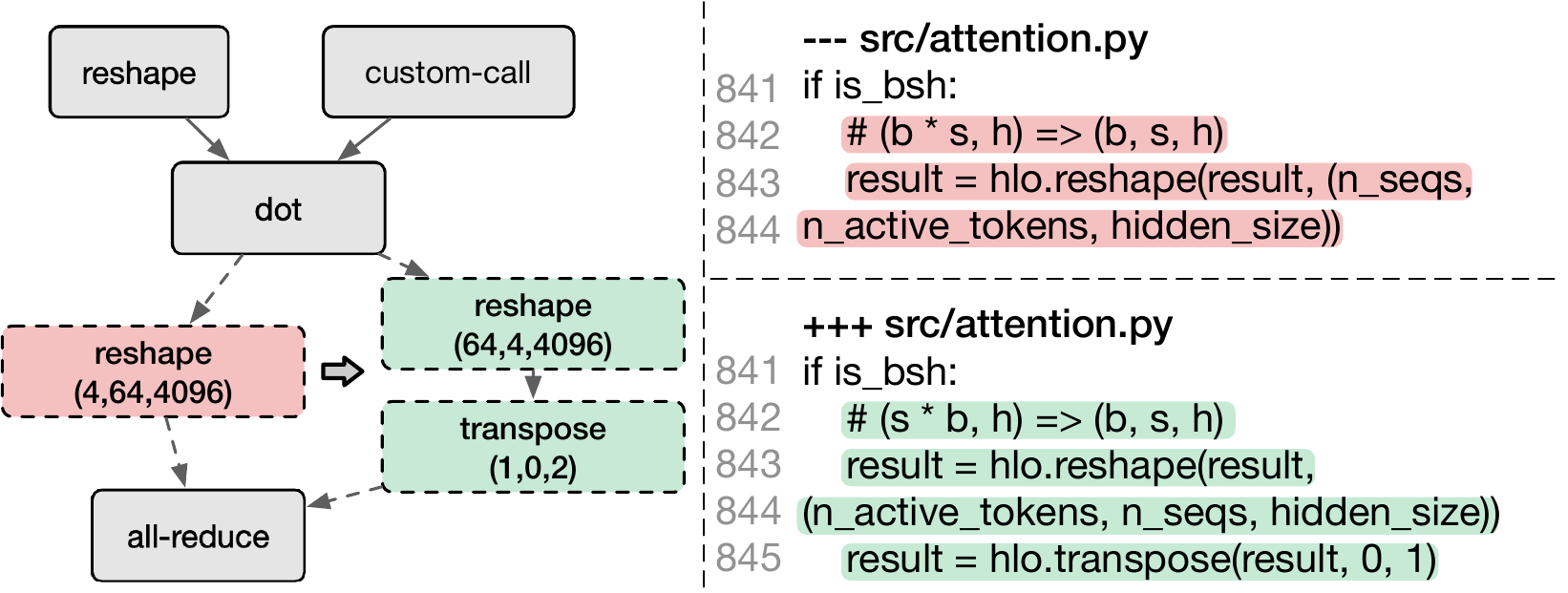}
   \caption{Incorrect layout transformation in BSH optimization.
   }
  \label{fig:BSH_bug}
\end{figure}

In this work we made a key observation that such silent errors are often 
reflected in non-equivalent semantics of the computational graphs. 
Deep neural networks (DNN) can be represented as a data flow graph (DFG), with nodes 
representing parameters, numerical or communication operators and the edges 
representing the flow of data.

We use a real-world bug found in {Transformers Neuronx} framework~\cite{transformersneuronx}, to demonstrate how to expose silent errors using computational graphs. 
Shown in Figure~\ref{fig:BSH_bug}, the bug manifest when using the Batch-Sequence-Head (BSH) layout for the output of the attention due to a wrong reshape operation \texttt{result} in the attention computation (the root cause is marked in red). The buggy version mistakenly reshapes the tensors to the BSH format \texttt{n\_seqs, n\_active\_tokens, hidden\_size}, while it should have accounted for the merging of axes from the \texttt{result} tensor, where the actual \texttt{result} tensor's dimensions is symbolized as \texttt{(s * b, h)}. The wrong reshape operation can lead to incorrect model outputs without explicit errors.
The fix was straightforward, requiring only a change in the reshape operation by switching the first and second dimensions, and then doing a transpose on those dimensions so that the final layout format is \texttt{n\_seqs, n\_active\_tokens, hidden\_size}. However, the debugging process took significant time due to the subtle nature of the reshape-transpose issue. Such issues could have been prevented if developers had  a solution for checking the semantic equivalence with respect to an oracle implementation at the computational graph level.

\setlength{\intextsep}{2pt}%
\setlength{\columnsep}{5pt}%
\begin{figure}[t]
    \centering
      \includegraphics[width=0.45\textwidth]{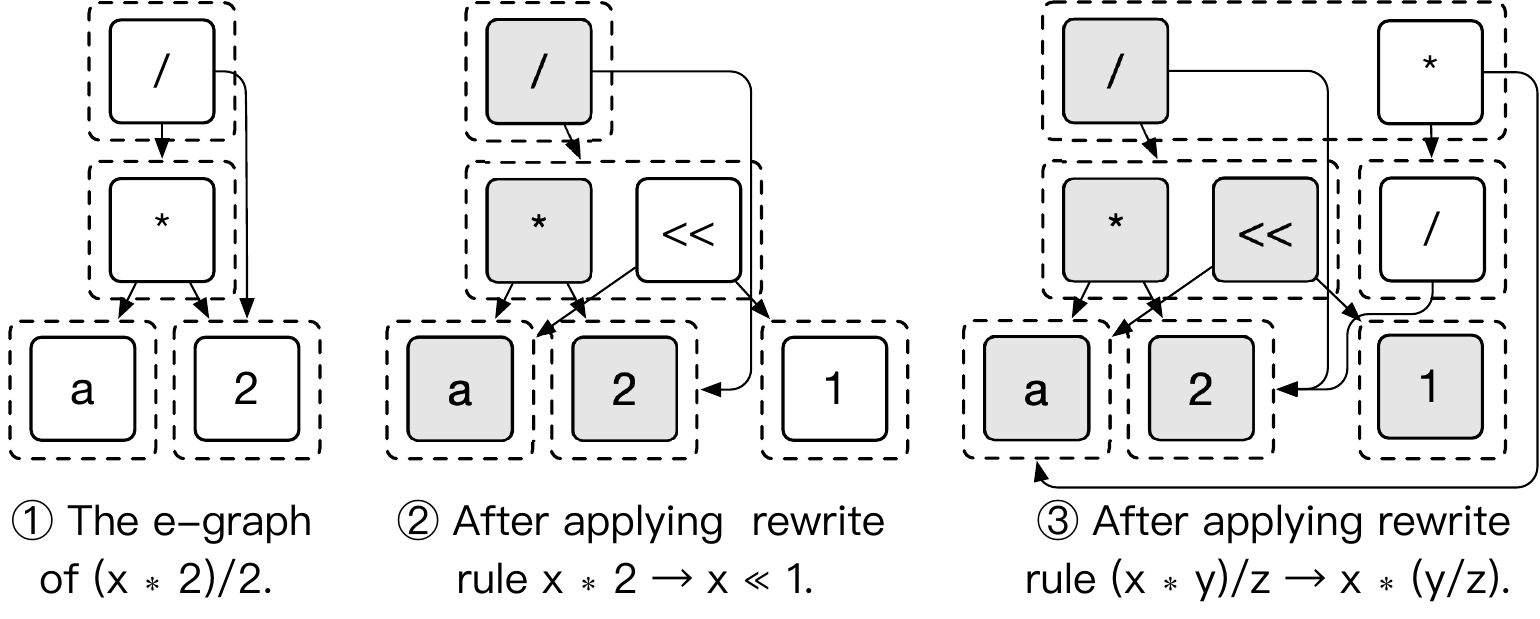}
   \caption{E-Graph example.
   }
  \label{fig:egraph}
\end{figure}

\subsection{Equality Saturation}

Equality saturation~\cite{es2009popl} is an emerging technique within compilers and programming languages.
It unifies all possible rewritten forms of an expression or program within a single, 
growing data structure called an e-graph. Rather than applying rewrite rules in a 
linear fashion, equality saturation systematically applies all valid rewrites in parallel,
making it an ideal option to explore multiple optimized  forms efficiently.

Figure~\ref{fig:egraph}, adapted from the egg paper~\cite{egg2021popl}, illustrates an 
example of an e-graph. An e-graph is a data structure used for representing and reasoning 
about program equivalences efficiently. It consists of e-classes (dashed boxes), which 
group structurally different but semantically equivalent expressions. Within each e-class 
are e-nodes (solid boxes), which represent individual expression terms.
Edges in the e-graph connect e-nodes to their respective child e-classes.

To use equality saturation, users usually provide a list of rewrite rules. 
Every rewrite rule has two components: an initial pattern and a transformed pattern. 
Both of these patterns are semantically equivalent, but they consist of different subgraphs. 
When rewrite rules are applied, new e-nodes and edges are 
introduced, but existing structures are never removed. Instead, newly added expressions 
are merged into the corresponding e-class, preserving equivalence relationships while 
expanding the set of recognized expressions.

In our work, we adopt egglog~\cite{egglog2023pldi} as the underlying e-graph engine, rather than using the original egg~\cite{egg2021popl}. While egg excels in compiler optimizations, our setting requires richer reasoning over tensor relations such as sharding, replication, and layout transformations, which are hard to express via graph-level transformations. Egglog extends e-graphs with a Datalog-style relational reasoning framework, enabling us to encode and propagate relations conveniently across tensors. %

\section{Approach: Verifying Equivalence with Equality Saturation}
\label{sec:approach}

\begin{figure}[t]
  \centering
    \includegraphics[width=0.48\textwidth]{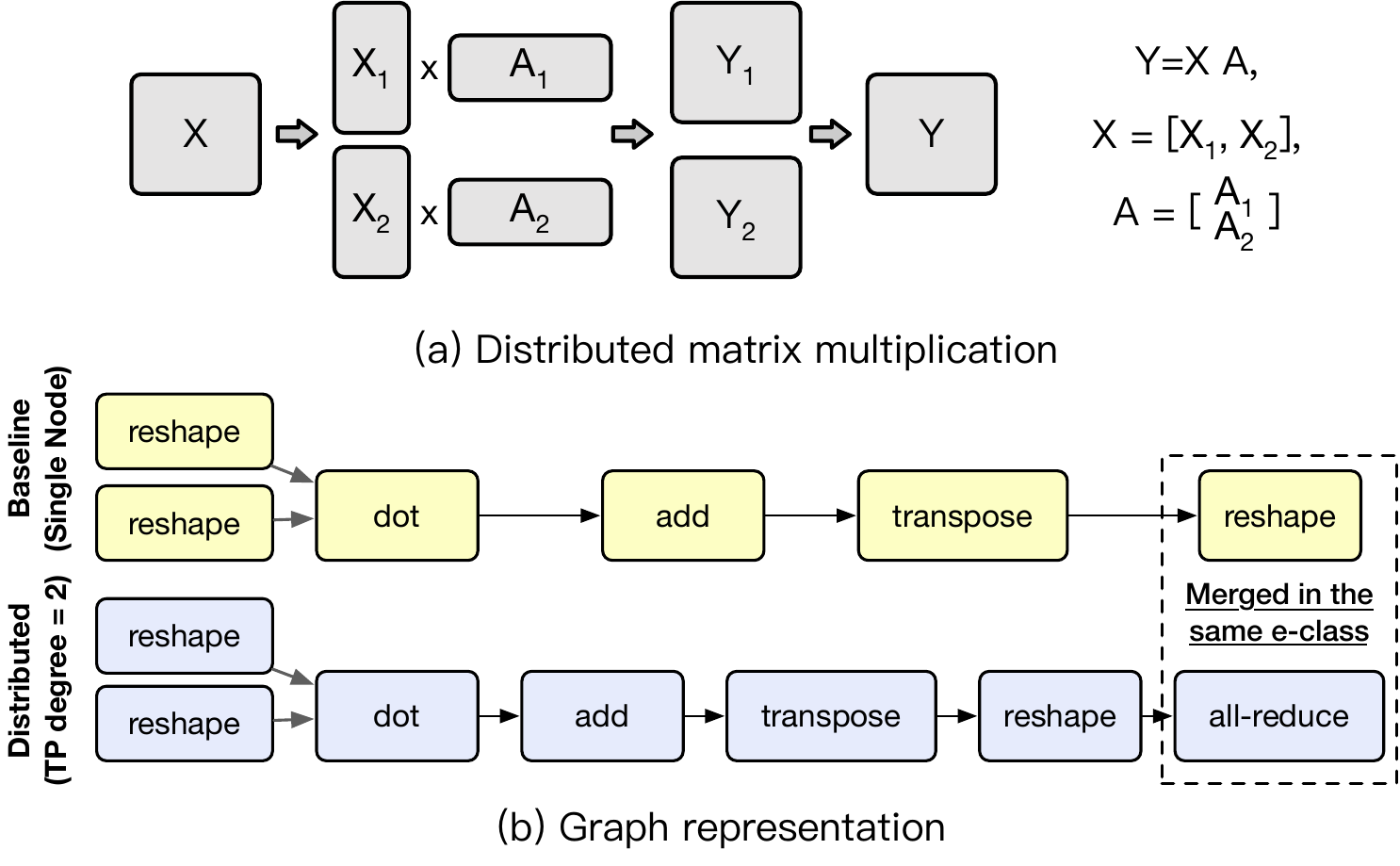}
  \vspace{-10pt}
 \caption{Matrix multiplication example. %
 }
\label{fig:example}
\end{figure}

We advocate an approach using equality saturation to verify graph equivalence to 
expose potential silent errors. 
Figure~\ref{fig:example} shows an example of checking correctness of  
matrix multiplication with two setups:
a single-node execution (top) 
and a tensor-parallel distributed version (bottom). 
The computation process includes 
matrix multiplication, addition, transposition, and reshaping. 
The distributed version partitions the input matrix 
across devices, computes local matrix multiplications independently, and then 
aggregates the results using \texttt{all-reduce}. 
Despite these structural difference, the final output tensors should remain 
semantically equivalent.

We first run the ML pipeline to 
generate two IR graphs: the original graph (in this case, with a single node)
and the transformed graph (with the distributed setup) using 
built-in ML functionalities. 
The original graph serves as the baseline version without complex transformations (thus more trustworthy),
and the transformed graph is a mutated version after enabling optimization features and/or adapted to 
distributed setups. The system records configurations for the transformed graph, 
including distribution degree and the model architecture.
Our goal is to construct a unified e-graph that integrates representations of both 
computational graphs, systematically identifying equivalent structures. 

We then iteratively expand the e-graph by incorporating nodes from IR graphs, 
applying rewrite rules, and merging nodes that represent the same structure
into equivalence classes (e-classes). 
We provide a list of rules that support common features such as tensor parallelism and 
developers can optionally add rules to support reasoning about new semantics.
As shown in the graph representation 
(b), both the single-node and distributed computations converge to the same 
e-class in the final stage. This means that even though the distributed version 
introduces additional steps, these transformations 
do not alter the final computation’s semantics. The e-graph captures all valid 
transformations, and the process stops once saturation is reached.

The two versions of models are semantically equivalent (verified)
if the output nodes of both graphs belong to the same e-class.
If developers observe severe performance degradation but the models are verified, 
this result eliminates the possibility of software bugs, which significantly reduces the search space of 
debugging (in this case, silent errors could be introduced from numerical errors 
or hardware faults). Otherwise, we will highlight dangling nodes that 
do not merge into a single equivalence class. Developers may further confirm 
this result by computing and comparing gradients to detect discrepancies in back propagation.

\section{Challenges in Verifying Large-Scale Models}
\label{sec:challenges}

Our approach demonstrates promise on the incorrect BSH case. However, we encounter several 
system challenges when 
applying our approach for larger and more structurally-different computation graphs.

\textbf{Computation cost explosion when e-graphs scale.} 
E-graph computation does not scale linearly with graph size.
We have observed the computation time and memory usage would dramatically increase if we naively apply e-graph analysis to complicated, multi-layer models.
Managing graph size to address this exponential growth 
is crucial for supporting deep learning models with hundreds of layers.

\textbf{Balancing rewriting rule generality and practicality.} 
Ideally the rewriting process should only leverage general rules to
capture more bug cases, but this creates a computation bottleneck—matching 
more nodes and including more patterns in the e-graph increases time and memory costs.
Meanwhile, highly specific rules cover fewer cases and 
bloat the rule set, and thus need to be manually revised for different workloads.

\textbf{Layout heterogeneity and semantic alignment.} 
Modern ML frameworks apply techniques like re-arrange, split/merge, shard/replicate, and partially reduce of tensors to enable large models and improve performance. Many distinct sequences of reshape/transpose yield tensors with the same shapes but not the same numerical content (the BSH bug in Figure~\ref{fig:BSH_bug}). 
Expressing in e-graph all possible sequences of equivalent layout transformations can be prohibitively expensive. 

\textbf{From equivalence verdicts to bug locations.} 
A binary ``unverified'' flag isn’t actionable. 
Large models often involve numerous  components and 
transformation steps -- manually searching  the whole code space is tedious. 
Developers need  guidance to resolve discrepancies beyond confirming bugs.

\section{System Design}
\label{sec:design}

\setlength{\intextsep}{2pt}%
\setlength{\columnsep}{5pt}%
\begin{figure}
    \centering
      \includegraphics[width=0.5\textwidth]{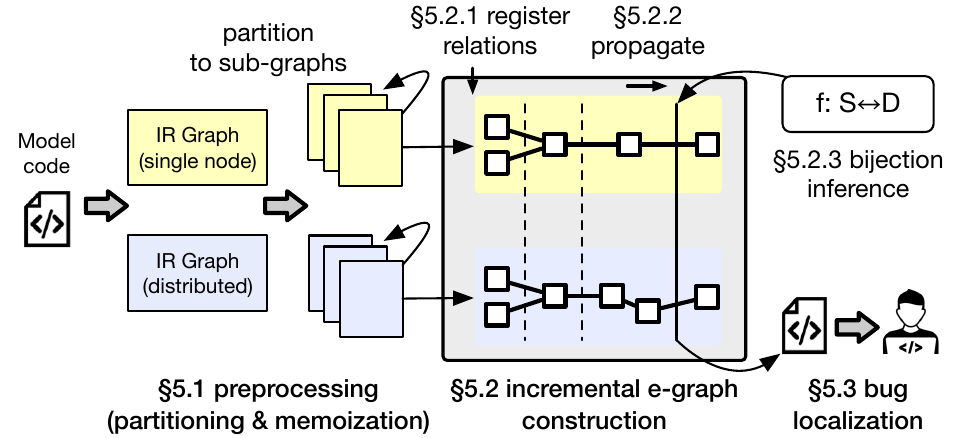}
   \caption{\sys workflow. 
   }
  \label{fig:workflow}
\end{figure}

We introduce \sys, a framework that automatically verifies the semantic equivalence 
of ML computational graphs from production-grade machine learning frameworks. %
\sys introduces the following key techniques:

 \begin{itemize}[noitemsep, topsep=0pt, partopsep=0pt, leftmargin=*]
  \item \textbf{Scaling e-graphs to production-scale computational graphs}: \sys partitions the 
  computational graphs into smaller subgraphs and performs parallel rewriting with layer memoization, enabling efficient, parallel processing without 
  overwhelming system resources (\S\ref{sec:design:partition}).
  \item \textbf{Layout and partition analysis via relational reasoning}: \sys employs 
  a scalable Datalog-style relation analysis to reason about the layout and partition of
  tensor slices onto different devices (\S\ref{sec:design:analysis}).
  \item \textbf{Bijection inference via symbolic execution}: When propagation alone is not enough to handle complex layout relations, we infer a bijection between baseline and distributed per-element tensors using symbolic execution. (\S\ref{sec:design:bijection}).
  \item \textbf{Discrepancy-based bug localization}: \sys not only determines verification outcomes but also traces discrepancies to their source code locations, enabling more efficient debugging (\S\ref{sec:design:mapping}).
\end{itemize}

\textbf{System Workflow.}
As shown in Figure~\ref{fig:workflow}, \sys follows a three-stage workflow. 
First, it generates single-device and distributed IR graphs using framework backends. 
Second, it preprocesses by \emph{partitioning} graphs by layers/topology and builds a unified e-graph: it registers input-tensor relations, traverses the distributed topology to propagate \emph{partition} and \emph{layout} facts, and performs \emph{bijection inference} (emitting the required reshape--transpose transforms). 
Finally, \sys halts at divergences and \emph{localizes bugs} by mapping discrepancies back to tensor-program source.

\subsection{Preprocessing: Graph Partitioning and Layer Memoization}
\label{sec:design:partition}

Before registering target IR graphs into the e-graph, \sys performs a preprocessing step. Large computational graphs can make rewriting prohibitively expensive, leading to exponential growth in both runtime and memory. To mitigate this scalability challenge, \sys introduces a two-step partitioning strategy that decomposes the input graph into smaller, more manageable subgraphs
as shown in Algorithm~\ref{algorithm:parallel_rewrite}.

\begin{figure}[t]
    \centering
      \includegraphics[width=0.45\textwidth]{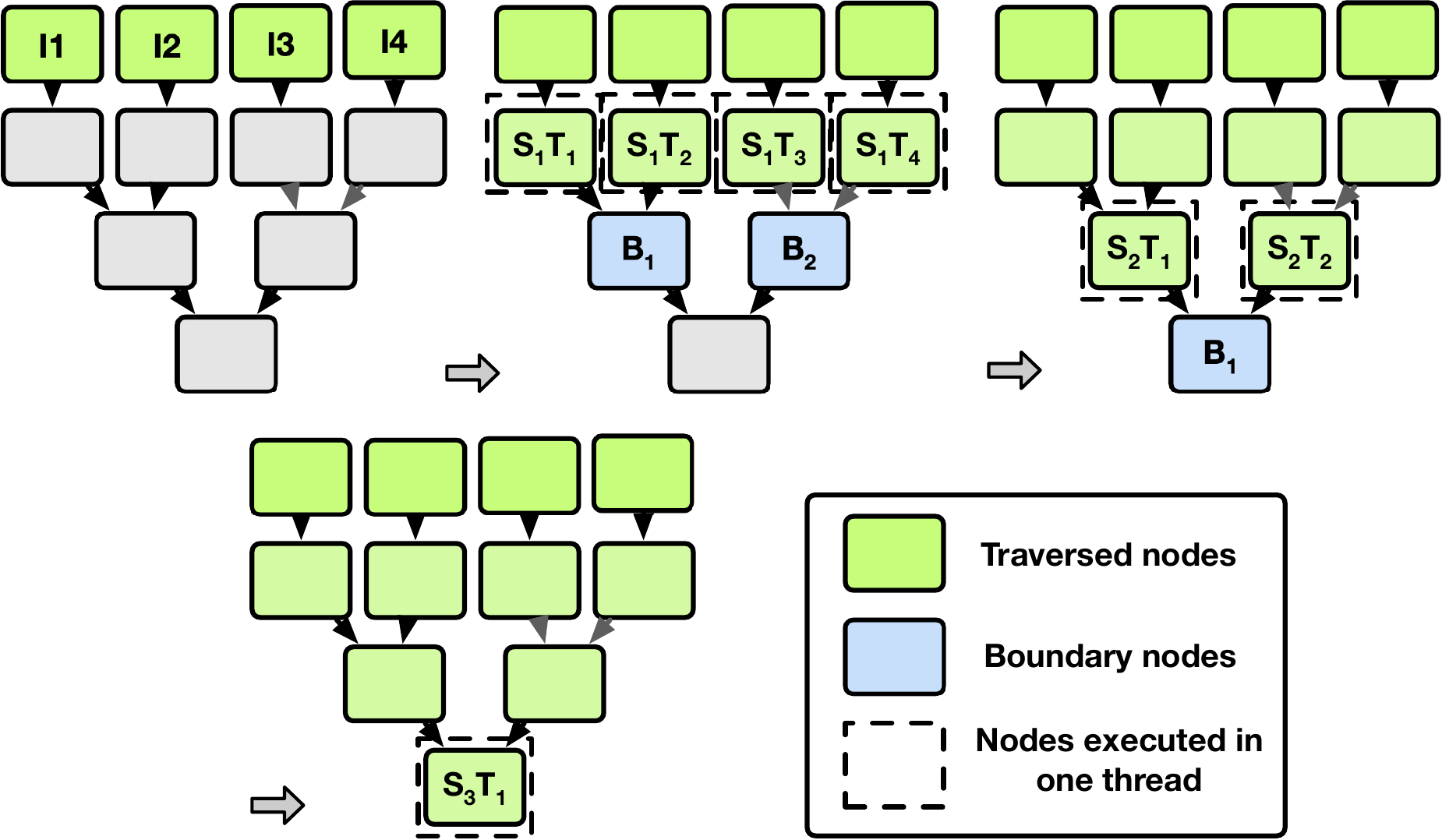}
   \caption{Illustration of the topological traversal. 
   }
  \label{fig:parallel_rewrite}
\end{figure}

\textbf{Partitioning.}
As a first step, \sys partitions the graph along neural network \textbf{layer boundaries}.
Layer boundaries serve as natural cut points: they preserve the semantic integrity of individual layers, align with the granularity of most framework and compiler optimizations~\cite{superoptimization2021mlsys}, and match the scope of vendor-provided kernels~\cite{tvm2018osdi}. This substantially reduces rewriting complexity while still capturing the majority of useful transformations.
Future extensions could explore cross-layer optimizations when cost-effective.
After per-layer partitioning, \sys further decomposes each subgraph in \textbf{topological} order. 
The key idea is to traverse from inputs and cut the graph at boundary nodes---nodes whose inputs are not yet fully registered in the e-graph.
This ensures that each subgraph is self-contained and can be processed independently.
As illustrated in Figure~\ref{fig:parallel_rewrite}, each stage  contains parallel topologies that can be rewritten concurrently.
Boundary nodes ($B_1, B_2$) delineate partitions, while groups such as $S_1T_1$ and $S_1T_2$ mark nodes in the same stage and executed in different rewriting threads.

\begin{algorithm}[tb]
\DontPrintSemicolon
\footnotesize
\caption{Two-step graph partitioning and parallel rewriting  
}
\KwIn{$G_s$ (single device), $G_m$ (multi device)}
\KwOut{$E$ (error code, if any)}
$(\mathcal{L}_s,\mathcal{L}_m)\leftarrow\textsc{PartitionGraphsToLayers}$($G_s$,$G_m$)\;
\ForEach(\tcp*[f]{layer pair}){$(L_s,L_m)$ in $(\mathcal{L}_s,\mathcal{L}_m)$}{
  \textsc{Register}($L_s$)\;
  \textsc{RegisterRelation}($(L_s.\mathit{in},L_m.\mathit{in})$)\;
  $\mathcal{S}\leftarrow$ \textsc{PartitionLayerToStages}($L_s$)\;
  \ForEach{stage $s$ in $\mathcal{S}$}{
     $\mathcal{T}\leftarrow$\textsc{InitThreads}($s.\mathit{topo}$)\;
     \tcp{update $(L_s.\mathit{out},L_m.\mathit{out})$}
     \textbf{parallel for all} $t\in\mathcal{T}$ \textbf{do}
        \textsc{Rewrite}($s.\mathit{topo}$[t])\; 
     \textbf{end parallel for}\;
  }
  $ok\leftarrow$ \textsc{Check}($(L_s.\mathit{out},L_m.\mathit{out})$)\;
  \lIf{$ok$}{\textsc{PropagateOutputToNextLayer}($(L_s.\mathit{out},L_m.\mathit{out})$)}
  \lElse{\Return \textbf{err\_abort}}
}
\label{algorithm:parallel_rewrite}
\end{algorithm}

\setlength{\intextsep}{2pt}%
\setlength{\columnsep}{5pt}%

Once the graphs are divided into different stages, \sys assigns multiple parallel threads for each stage
(\emph{e.g.,} $T_1$--$T_4$ in Figure~\ref{fig:parallel_rewrite}). 
Correctness is preserved because the relations computed in each thread depend only on nodes whose inputs have already been analyzed in earlier steps. Dependencies that cross subgraphs are handled naturally by boundary nodes: a boundary node is only rewritten in a later stage once all of its inputs have been resolved in previous stages.

\textbf{Layer memoization.}
Large models often contain layers with highly similar or identical structures.
Rewriting every such layer independently would repeat the same analysis. To avoid this redundancy,
\sys memoizes the results of previously rewritten subgraphs. Each partitioned subgraph is assigned a fingerprint derived from its single-device and distributed forms. When a new subgraph shares the same fingerprint as an earlier one, \sys reuses the memoized relations instead of re-executing the rewriting. Our experience shows this significantly reduces duplicate work across layers.
\textbf{Soundness.}
The aforementioned partitioning and memoizations techniques preserve soundness of \sys as a verification tool.
It is possible for \sys to miss rewrite opportunities and relations across partition boundaries such that certain
correct computational graphs cannot be verified, also to re-verify layers that are structurally different but semantically equivalent. However, soundness of \sys is preserved as the
relation analysis and e-graph rewrite rules are sound.

\subsection{Processing: Incremental E-Graph Construction}

Next 
we discuss how semantic relations between single-device and distributed graphs are actually established.
Simply applying e-graph rewriting is insufficient: many operators reshape, shard, or merge tensors in ways that require explicit reasoning about their relations.
To handle this, \sys incrementally constructs the e-graph, adding new nodes and relations step by step as it traverses the computation graphs. Each stage of this process activates  relation reasoning---starting from input registration, then propagating relations through operators, and extending propagation with bijection inference when the two graphs differ structurally.

\setlength{\intextsep}{2pt}%
\setlength{\columnsep}{5pt}%
\begin{figure*}
    \centering
      \includegraphics[width=\textwidth]{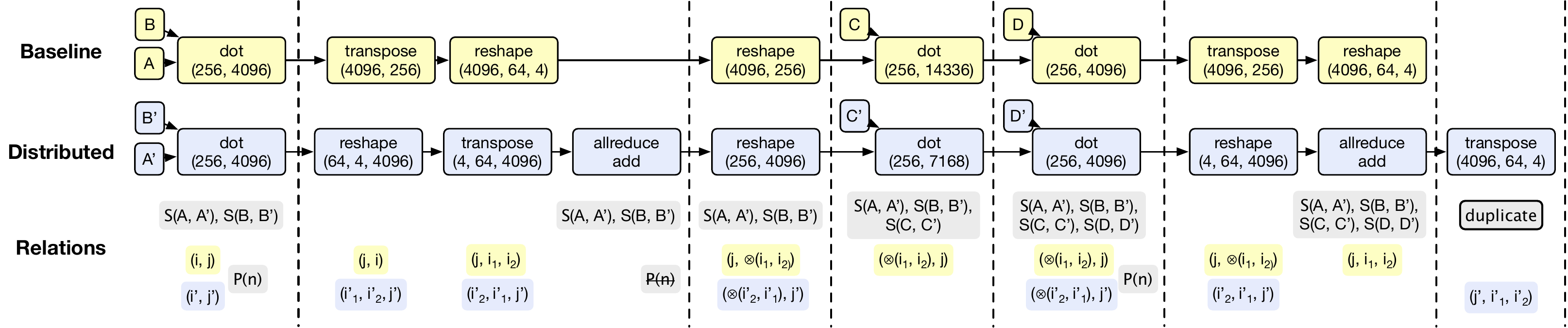}
   \caption{A full computation graph of different reshape, transpose and dot operators with distributed semantics. We show how the  partition and layout analysis is integrated with the propagation process (via rule templates defined from Table \ref{tab:rule_templates}).
   \textit{S} represents shard relations.
   \textit{P} represents layout relation of partial tensors.
   $\otimes$ represents combination of axes. %
   }
  \label{fig:axes_mapping_report}
\end{figure*}

\subsubsection{Registering Input Relations}
\label{sec:design:input_relations}

The generated graphs from ML compilers by default do not contain information on how the input tensors are related to one another. 
\sys solves this issue by creating relations between graphs by instrumenting compilers and logging sharding/replication information during the IR generation phase. The information is captured as sharding/replication annotations, \emph{e.g.,} \texttt{shard\_along(self, tensor, dim)}. %
These annotations also show the device placement of the distributed tensors.

Certain auxiliary tensors such as those encoding device IDs, \emph{e.g.,} \texttt{torch.arange(tp\_degree), sharding=0} are not automatically annotated. For these types of information, \sys supplements the missing information by manually specifying which tensors carry device metadata and how they should be interpreted. 
These registered input relations serve as the foundation for subsequent reasoning. We discuss how to formally encode these relations in the following subsection.

\subsubsection{Propagating Relations with Layout and Partition Analysis}
\label{sec:design:analysis}

Once input relations are registered, \sys incrementally propagates them through intermediate operators to determine whether the two graphs remain equivalent. This propagation combines several complementary analyses, each addressing different gaps.

\begin{figure}[t]
\label{fig:syntax-semantics}
\scriptsize

\begin{bnf}(
)[
colspec = {l@{}l@{}c@{}l@{}l},
colsep = 1pt,
]
$t$ : \textsf{Tensor} ::=
| $x, y, z$ : baseline tensors
| $x', y', z'$ : distributed tensors  
| \texttt{$\mathit{op}$($t_1$, \ldots, $t_n$)} : operations
;;
$\ell$ : \textsf{Layout} ::=
| \textit{transpose($\pi$)} : transpose with permutation $\pi$
| \textit{reshape($s$)} : reshape to shape $s$
| $\ell_1 \circ \ell_2$ : layout composition
| \textit{bijection($s_1$, $\pi$, $s_2$)} : abbr. for reshape-transpose-reshape
;;
$A$: \textsf{SymAxis} ::=
| $i, j, \dots$ : baseline tensor symbolic axes
| $i', j', \dots$ : distributed tensor symbolic axes
;;
$\mathit{AE}$: \textsf{AxisExp} ::=
| $A$ :  symbolic axis
| $\otimes\left(\mathit{AE}, \mathit{AE}\right) $ : combination of axes
;;
$M$: \textsf{AxisMap} ::=
| $\emptyset$ : empty axis map
| $M \cup \{AE \leftrightarrow AE\}$ : axis expression correspondence
;;
$\mathit{op}$ : \textsf{BaseOp} ::=
| \textit{elem\_op} : element-wise operation
| \textit{dot} : matrix/vector multiplication
| \textit{max\_reduce(d)} : max reduction along dim d
| \textit{sum\_reduce(d)} : sum reduction along dim d
| \textit{apply\_layout}$\left(\ell \right)$ : apply layout transformation
| \textit{slice(d, i, l)} : $l$-slice along dim $d$ at index $i$
;;
$\mathit{op}'$ : \textsf{DistOp} ::=
| \textit{$\mathit{op}$'} : dist. version of base op
| \textit{all-reduce'($c$, $\mathit{op}$)} : dist. all-reduce along $c$ cores
| \textit{all-gather'($d$, $c$)} : dist. all-gather along $c$ cores at dim $d$ 
| \textit{reduce-scatter'($d$, $c$, $\mathit{op}$)} : dist. reduce-scatter along $c$ cores at dim $d$ 
;;
$R$ : \textsf{Rel} ::=
| \texttt{sharded($t$, $t'$, $d$, $c$)} : $t'$ is $t$ partitioned along dim $d$ on $c$ cores
| \texttt{duplicate($t$, $t'$, $c$)} : $t'$ is an identical replica of $t$ on $c$ cores
| \texttt{layout($t$, $t'$, $\ell$)} : $t'$ can be transformed to $t$ using layout $\ell$
| \texttt{partial($t$, $t'$, $c$, $\mathit{op}$)} : partial result across $c$ cores to be reduced
| \texttt{slice($t$, $t'$, $r$, $b$, $d$,} : $t' = t$ is a slice of $b$ along dim $d$ at index $x$\\
& & & \hspace{5mm}\texttt{ $x$, $l$)} &  with length $l$, $t'$ lives on rank $r$
| \texttt{loop\_red\_B($\mathit{op}$, $d$, $t$, } : $t$ is partial reduction $\mathit{op}$ on $b$ on dim $d$ \\
& & & \hspace{6mm}\texttt{$ts$, $b$)}&  using slices $\mathit{ts}$   
| \texttt{loop\_red\_D($\mathit{op}$, $d$, $t'$, } : $t'$ is partial reduction $\mathit{op}$ on $b$ on dim $d$\\
& & & \hspace{6mm}\texttt{$ts$, $b$, $r$)}&   of baseline slices $\mathit{ts}$, $t'$ lives on rank $r$
\end{bnf}
\caption{Syntax of relational analysis. We use italics to denote operations and monospace to denote relation symbols.
}
\label{fig:relations} 
\end{figure}

Figure~\ref{fig:relations} gives the {relational language} that \sys\ reasons about.
It separates (i) \emph{objects} from (ii) \emph{operators} and (iii) \emph{relations}.
Objects include baseline and distributed tensors ($t$ vs.\ $t'$) and layout terms $\ell$ (compose, reshape, transpose, and the $\textit{bijection}(s_1,\pi,s_2)$ for a reshape–transpose–reshape triple).
Axes are symbolic ($i,j,\ldots$) and may combine via $\otimes$ to describe mesh products; $M$ records an axis correspondence between baseline and distributed views.
Operators summarize only what the analysis needs: element-wise ops, dot, layout application, slicing, and three collectives (all-reduce, all-gather, reduce-scatter).
Relations are the facts \sys\ maintains/propagates: \texttt{sharded} capture data placement; \texttt{layout} records a provably bijective layout transform; \texttt{partial} and the looped \texttt{loop\_red\_B}/\texttt{loop\_red\_D} track per-core partial results that must be discharged by a reduction or collective operation.

Table~\ref{tab:rule_templates} lists the Datalog-style rules that drive our analysis.
Rules fall into five families: Partition (propagate sharded), Layout (synthesize/check bijective layouts), Slicing (finer-grained partition), Unroll (handle \texttt{loop\_red\_B}/\texttt{loop\_red\_D}), and a mix of categories above.
We write Datalog-style rules as $H \leftarrow B$ (``if $B$ holds, derive $H$''). Formally, $B_1$, …, $B_n$ means conjunction and the rule denotes ($B_1 \wedge … \wedge B_n$).

\begin{table*}[t]
\centering
\footnotesize
\caption{A subset of Datalog-style analysis rules for partition and layout analysis following the syntax in Figure~\ref{fig:relations}. We additionally use $x \leadsto w$ to mean the sequence of layout transformations. In the actual implementation of \sys, many rules are  polymorphic over operator types.}
\begin{tabular}{l l}
\toprule
\textbf{Rule} & \textbf{Type} \\ 
\midrule

$\texttt{sharded}\left(\textit{elem\_op}\left( x\right), \textit{elem\_op'}\left( x' \right), d, c\right) \leftarrow \texttt{sharded}\left(x, x', d, c\right), \textit{elem\_op}\left(x\right), \textit{elem\_op'}\left(x'\right)$ & Partition \\[0.3em]

$\texttt{partial}\left(z, z', c, \textit{add}\right) \leftarrow \texttt{sharded}\left(x, x', d, c\right), \texttt{sharded}\left(y, y', d, c\right), z=\textit{dot}\left(x, y\right), z'=\textit{dot'}\left(x', y'\right)$ & Partition + Layout \\[0.3em]

$
\texttt{sharded}\left(z, z', d, c\right), 
\texttt{layout}\left(z, z', \ell', c\right),
\ell'  = \textsc{bijection\_inference}\left(\ell, x \leadsto w , x' \leadsto w' \right)$ & \multirow{2}{*}{Layout} \\ \hspace{3mm}
$\leftarrow 
\texttt{layout}\left(x, x', \ell, c\right),
\texttt{sharded}\left(y, y', d, c\right),
z=\textit{dot}\left(w, y\right), 
z'=\textit{dot'}\left(w', y'\right)$ &  \\[0.3em]

$\texttt{layout}\left( 
    z, 
    x', 
    \ell \circ \textit{transpose}\left(\pi\right) , 
    c
    \right) 
\leftarrow \texttt{layout}\left( x, x', \ell , c\right), z =\textit{transpose}\left(x, \pi\right)$ & Layout \\[0.3em]

$\texttt{layout}\left( 
    x, 
    z', 
   \textit{reshape}\left(x'.\textit{shape}\right) \circ \ell  , c\right) 
\leftarrow \texttt{layout}\left( x, x', \ell, c \right),
z'=\textit{reshape}\left(x', s'\right)$
& Layout \\[0.3em]

$\texttt{layout}\left(p, q', \ell', c\right),
\ell' =  
\textsc{bijection\_inference}\left(\ell, x \leadsto p, z' \leadsto q' \right)$ & \multirow{2}{*}{Partition + Layout}\\ \hspace{3mm}
$\leftarrow \texttt{layout}\left(x, x', \ell, c\right), z'=\textit{all-reduce'}\left(x', c, \mathit{add}\right),
\texttt{partial}\left(x, x', c, \textit{add}\right) $ &  \\[0.3em]

$\texttt{layout}\left(p, q', \ell', c\right),
\ell' = \textsc{bijection\_inference}\left( \ell, x \leadsto p, z' \leadsto q' \right) $ & \multirow{2}{*}{Partition + Layout}\\ \hspace{3mm}
$\leftarrow \texttt{layout}\left(x, x', \ell, c\right), z' = \textit{reduce-scatter'}\left(x', d, c, \mathit{add}\right),
\texttt{partial}\left(x, x', c, \textit{add-concat}\right) $ &  \\[0.3em]

$\texttt{duplicate}\left(x, x', c\right) \leftarrow \texttt{sharded}\left(x, x', d, c\right), \textit{all-gather'}\left(x', d, c\right)$ & Partition \\[0.3em]

$\texttt{duplicate}\left(x, z', c\right) \leftarrow \texttt{layout}\left(x, x', \ell, c\right), z'=\textit{apply\_layout'}\left(x', \ell\right)$ & Layout \\[0.3em]

$\texttt{partial}\left(z, z', c, \textit{max}\right) \leftarrow \texttt{sharded}\left(x, x', d, c\right), z=\textit{max\_reduce}\left(x, d\right), z'=\textit{max\_reduce'}\left(x', d\right)$ & Partition + Layout \\[0.3em]

$\texttt{slice}\left(z, z', r, b, d, j, l\right) 
\leftarrow 
\texttt{sharded}\left( x, x', d, c\right),
z = \textit{slice}\left(x, d, j, l\right),
z' = \textit{slice}\left(x', d, k, l\right),
\texttt{rank}\left(z'\right) = r,
k = rl
$ & Slicing \\[0.3em]

$\texttt{loop\_red\_B}\left( \mathit{add}, d, x, \emptyset, b \right)
\leftarrow
\texttt{slice}\left(x, x', r, b, d, 0, l\right)
$ & Slicing + Unroll \\[0.3em]

$\texttt{loop\_red\_B}\left( \mathit{add}, d, z, xs \cup \{ x_1 \}, b \right)
\leftarrow
z = \mathit{add}(x, x_1),
\texttt{loop\_red\_B}\left( \mathit{add}, d, x, xs , b \right),
\texttt{slice}\left(x_1, x_1', r, b, d, k, l\right)
$ & Slicing + Unroll \\[0.3em]

$\texttt{loop\_red\_D}\left( \mathit{add}, d, x', \emptyset, b, r\right)
\leftarrow
\texttt{slice}\left(x, x', r, b, d, 0, l\right)
$ & Slicing + Unroll \\[0.3em]

$\texttt{loop\_red\_D}\left( \mathit{add}, d, z', xs \cup \{ x_1' \}, b, r \right)
\leftarrow
z' = \mathit{add}(x', x_1'),
\texttt{loop\_red\_D}\left( \mathit{add}, d, x', xs, b \right),
\texttt{slice}\left(x_1, x_1', r, b, d, k, l\right)
$ & Slicing + Unroll \\[0.3em]

$\texttt{loop\_red\_D}\left( \mathit{add}, d, z', \bigcup_r xs_r, b \right)
\leftarrow
z'=\textit{all-reduce'}\left(x', c,\mathit{add}\right),
\forall r. \texttt{loop\_red\_D}\left( \mathit{add}, d, x', xs_r, b, r \right)
$ & Unroll \\[0.3em]

$\texttt{duplicate}\left(z, z', c\right) \leftarrow
\texttt{loop\_red\_B}\left( \mathit{add}, d, z, xs, b \right),
\texttt{loop\_red\_D}\left( \mathit{add}, d, z', xs, b, r \right),
b = \mathit{cat}((xs_1, \dots, xs_n), \textit{dim}=d)
$ & Unroll \\[0.3em]

\bottomrule
\end{tabular}
\label{tab:rule_templates}
\end{table*}

\textbf{Partition and layout analysis.}
Partition analysis handles relations linking a baseline tensor \(t\) with its distributed counterpart \(t'\) and records placement, \emph{e.g.,} \emph{sharded} along dimension \(d\) across \(c\) devices. Layout analysis tracks how a \emph{layout} relation \(\texttt{layout}(t,t',\ell)\), which states that \(t\) and \(t'\) are connected by a \emph{bijective} layout \(\ell\).

Here we walkthrough Figure~\ref{fig:axes_mapping_report} to demonstrate how relations propagate in the constructed e-graph.
The top row is the baseline graph (yellow), the middle row the distributed graph (blue), and the bottom row the derived relations.
It starts with the sharded relation between $A$ and $A'$
as well as  $B$ and $B'$.
Because the contracted dimension is sharded, Table Table~\ref{tab:rule_templates}’s partial-from-reduction rule fires, and produces a partial relation $P(n)$, where n stands for the number of cores. The tool symbolizes the shape of tensors to be $(i, j)$ and $(i', j')$. Next,
baseline applies transpose then reshape; the distributed side performs the dual sequence plus allreduce. We align the two paths by inferring a bijective layout (we explain in more details in Section~\ref{sec:design:bijection}),
which yields an axis correspondence $(i_2', i_1', j') \leftrightarrow (j, i_1, i_2)$. The allreduce node 
eliminates the partial relation and restores a normal tensor relation according to the collective discharge rule.
The subsequent steps repeat similar patterns. The final layout step plus collective yields a duplicate output whose axes are aligned (rightmost column).

\setlength{\intextsep}{2pt}%
\setlength{\columnsep}{5pt}%
\begin{figure}
    \centering
      \includegraphics[width=0.37\textwidth]{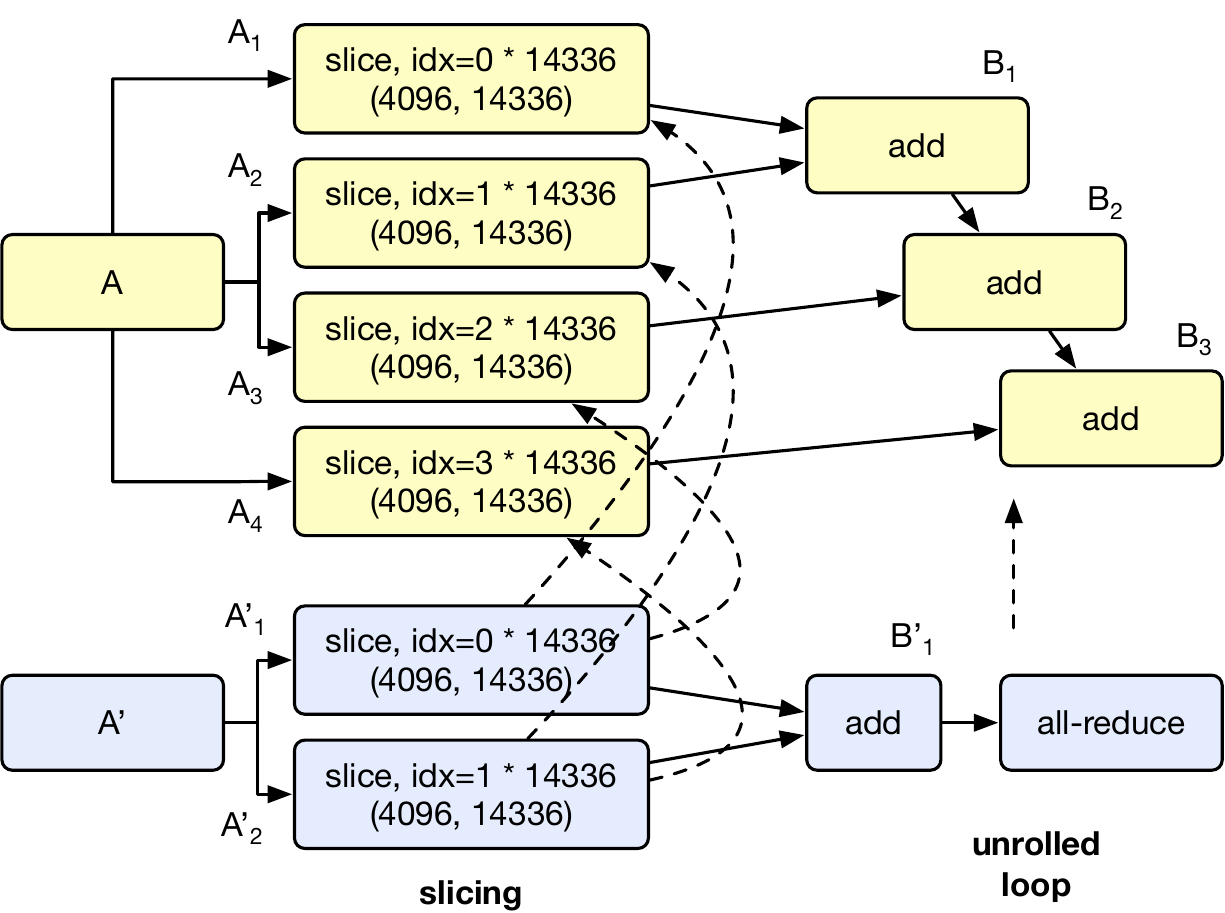}
   \caption{A computation graph with fine-grained slicing and unroll loop analysis. %
   }
  \label{fig:slice_reduce_relations}
\end{figure}

\textbf{Fine-grained slicing.}
Some parallelization strategies introduce slicing patterns that cannot be fully captured by partition analysis. 
For these fine-grained slicing operations, \sys utilizes slice relations to relate specific distributed tensors to the single-device versions. 
This case arises when simply concatenating the distributed slices reconstructs the baseline tensor. 
For example, Figure \ref{fig:slice_reduce_relations} illustrates how fine-grained slicing produces equally sized sub-tensors whose equivalence must be inferred explicitly (tensor $A_1$ is equivalent to tensor $A_1$ and $A'_1$ is also equivalent to tensor $A_3$ at different devices). 
Slice analysis can be viewed as a special case of partition analysis, extending it to finer levels of granularity.

\textbf{Unroll loop analysis.}
Some distributed parallelization strategies unroll recursive element-wise operations into loops, thus it would require an unbounded set of rules if hard-coding the operator count. %
To address this, \sys introduces \emph{reduce relations}, which relate a distributed tensor to the collection of partial tensors produced in the baseline graph. 
As illustrated in Figure~\ref{fig:slice_reduce_relations}, although tensors $B'_1$ and $B_1$ share the same shape, they are not equivalent. Instead, $B'_1$ corresponds to the reduction of multiple baseline tensors ($B_1$--$B_3$). 
By representing these equivalences as reduce relations, \sys can correctly reason about unrolled loops and recursive elementwise computations.

\setlength{\intextsep}{2pt}%
\setlength{\columnsep}{5pt}%
\begin{figure*}[ht]
    \centering
      \includegraphics[width=0.95\textwidth]{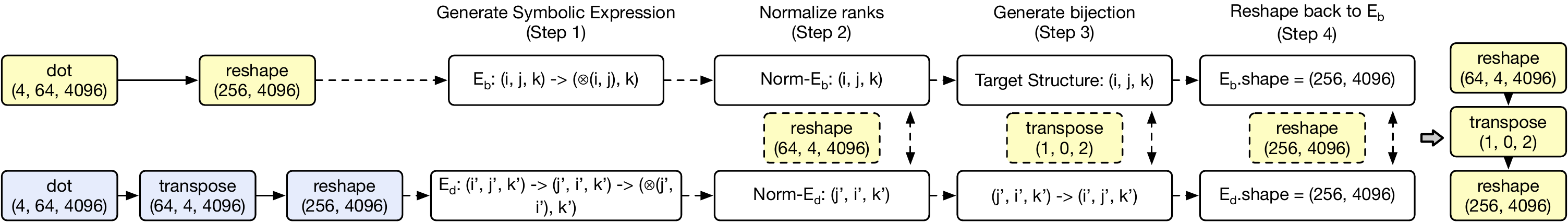}
   \caption{The bijection inference process described in Algorithm \ref{algo:infer_bijection_expressions}. 
   }
  \label{fig:bijection_case_study_update}
\end{figure*}

\subsubsection{Extending Relation Propagation with Bijection Inference}
\label{sec:design:bijection}

The propagation of layout relations becomes especially challenging when both graphs are consumed by different reshape-transpose sequences. For example, relating \texttt{reshape(transpose(x))} with \texttt{transpose(y)} is difficult if \sys uses relation propagation because the structure between them is different.
Encoding rules to relate different and equivalent layout transformations is infeasible due to the huge number of possible layout transformations.
\sys resolves this issue by inferring a bijection that relates two different layout transformations with each other.  
\sys aims to relate two layout transformations because many optimization techniques encode different layout transformations, and ensuring the equivalence of them is crucial to maintain correctness. The bijection inference checks whether the different layout transformations are indeed equivalent to each other. If not, \sys generates a concrete sequence of reshape-transpose sequence that converts the layout of the distributed tensor to be equivalent to the baseline tensor, and \sys  checks whether this bijection is defined later in the computation graphs.

\textbf{Inference algorithm.}
Algorithm~\ref{algo:infer_bijection_expressions} shows how \sys calculates the bijection between different layout transformations. 
We model each path’s layout sequence as symbolic axis expressions \(E_b\) and \(E_d\) (Step 1).
As shown in Figure \ref{fig:bijection_case_study_update}, \sys symbolizes shapes (4, 64, 4096) as $(i, j, k)$, changing it to $(\otimes(i, j), k)$ when consumed by \texttt{reshape} to reflect the merging of axes. \sys normalizes the ranks so that they have the same rank by merging or splitting (Step 2). 
\sys then computes the permutation between $\hat{E}_{\text{d}}$ and $\hat{E}_{\text{b}}$ by finding index $j$ in $\hat{E}_{\text{d}}$ such that it is structurally equal to index $i$ in $\hat{E}_{\text{b}}$ (Step 3).
 \sys converts the distributed axes map $(j', i', k')$ into $(i', j', k')$ because it has the same structure as the baseline axes map $(i, j, k)$. The generated permutation indices is (1, 0, 2), which is represented as a transpose operation. Finally, \sys constructs the final operation sequence in Step 4. \sys adds a final reshape operation to $Ops$ so that $E_{\text{b}}$ has same shape as $E_{\text{d}}$.  $E_{\text{d}}$ is reshaped to (256, 4096) so that it has same shape as $E_{\text{b}}$. The generated bijection is \texttt{[reshape(64, 4, 4096), transpose(1, 0, 2), reshape(256, 4096)]}. \sys checks whether this bijection makes the distributed layout sequence equivalent to the baseline version. If not, \sys returns None.

\sys limits symbolic execution to structurally different layout transformations, unlike TrainVerify~\cite{trainverifySOSP2025} which applies it to entire graphs. This narrower scope avoids unnecessary heavy reasoning for operators that don’t require it.

 \begin{algorithm}[t]
\footnotesize
\caption{Inferring bijection mapping}
\label{algo:infer_bijection_expressions}
\KwIn{
  $\ell$: existing layout relation between baseline tensor \texttt{b} and distributed tensor \texttt{d};\\ 
  $S_{\text{b}}$: list of layout ops on baseline path;\\
  $S_{\text{d}}$: list of layout ops on distributed path.
}
\KwOut{
  A $\textit{bijection}\left(s_1, p, s_2\right)$ that maps \texttt{d} to \texttt{b},
  or $\bot$ if no bijection exists.
}

\BlankLine
\tcp{Step 1: Generate symbolic expression and mapping between symbolic axes of \texttt{b} and \texttt{d}}
$M \gets \mathrm{ExtractAxisMap}\left( \ell, b.\mathit{shape}, d.\mathit{shape} \right)$,
$E_{\text{b}} \leftarrow \mathrm{GenExp}\left(S_{\text{b}}\right)$, 
$E_{\text{d}} \leftarrow \mathrm{GenExp}\left(S_{\text{d}}\right)$

\BlankLine
\tcp{Step 2: Rank normalization}
$(\hat{E}_{\text{b}}, \hat{E}_{\text{d}}, impossible) \leftarrow \mathrm{NormalizeRank}\left(E_{\text{b}}, E_{\text{d}}\right)$ \\
  \lIf{impossible}{\Return $\bot$}

\BlankLine
\tcp{Step 3: Find permutation bijection}
$p \leftarrow$ an empty permutation of length $\mathrm{rank}\left(\hat{E}_{\text{b}}\right)$ \\
\For{$i \gets 0 ~\KwTo~ \mathrm{rank}\left(\hat{E}_{\text{b}}-1\right)$}{
  find $j$ such that $\hat{E}_{\text{d}}\left[j\right]$ is equal to $\hat{E}_{\text{b}}\left[i\right]$ under $M$ \\
  \lIf{no such $j$ or $j$ reused}{\Return $\bot$}
  $p\left[i\right] \leftarrow j$
}

\BlankLine
\tcp{Step 4: Construct the operation sequence}
$\mathsf{Ops} \leftarrow $ [] \\
\If{$\hat{E}_{\text{d}} \neq {{E}_\text{d}} \vee rank\left(\hat{E}_{\text{d}}\right) \neq rank\left(\hat{E}_{\text{b}}\right)$}{
  append $\mathrm{reshape}\left(\cdot \rightarrow shape_{normalized}\right)$ to $\mathsf{Ops}$
}
append $\mathrm{transpose}\left(p\right)$ to $\mathsf{Ops}$ \\
\If{$\hat{E}_{\text{d}}.shape \neq {{E}_\text{b}}.shape$}{
append $\mathrm{reshape}\left(\cdot \rightarrow {E}_\text{b}.shape\right)$ to $\mathsf{Ops}$
}

\Return $\mathsf{Ops}$
\end{algorithm}

\textbf{Scope assumptions. }
The numerous equivalent and different sequence layout transformations make it computationally challenging to infer a bijection mapping that converts one sequence into another.
\sys limits the scope of the algorithm to reshape operations that merges or split dimensions within a tensor. This is because we find that many production-grade ML frameworks reshape the tensors on a grouping mechanism. Figure \ref{fig:axes_mapping_report} has several examples where the reshape operation is either a splitting or merging of dimensions (\emph{e.g.,} reshape from (256, 4096) to (64, 4, 4096)).

\subsection{Postprocessing: Localizing Code Bugs based on E-Graph Discrepancies}
\label{sec:design:mapping}

Our tool 
should not only give the binary result (verified/unverified) but also aid developers 
in locating the bugs. 
To facilitate this, \sys instruments the ML compiler using the tool logging API 
to capture source-level 
debugging information during graph generation. 
\sys stores metadata
and associates source AST nodes with IR graph nodes when constructing the IR graphs from the source code.
Each IR graph node is extended with a unique identifier,
which points to the source file, function, and line number,
\emph{e.g.,} \texttt{\{source line: ``flash\_decoding.py:42'', expr: "hlo.exp(...)"\}},
which is preserved at nodes when converting IR graphs to e-graphs.

During the rewrite process, \sys checks whether the node \sys rewrites belongs to a relation or not, and stores each node into two categories: verified and unverified.  
Generating a list of unverified nodes is not helpful since in non-equivalent graphs, the list of unverified nodes can be numerous, and the source of the issue is not the unverified node. 
When \sys detects non-equivalence, 
\sys goes through every unverified node and checks whether the inputs of that node is verified or not.
\sys generates unverified nodes where the inputs are verified, also with the corresponding source line and file number.  

\setlength{\intextsep}{2pt}%
\setlength{\columnsep}{5pt}%
\begin{figure}[t]
    \centering
      \includegraphics[width=0.48\textwidth]{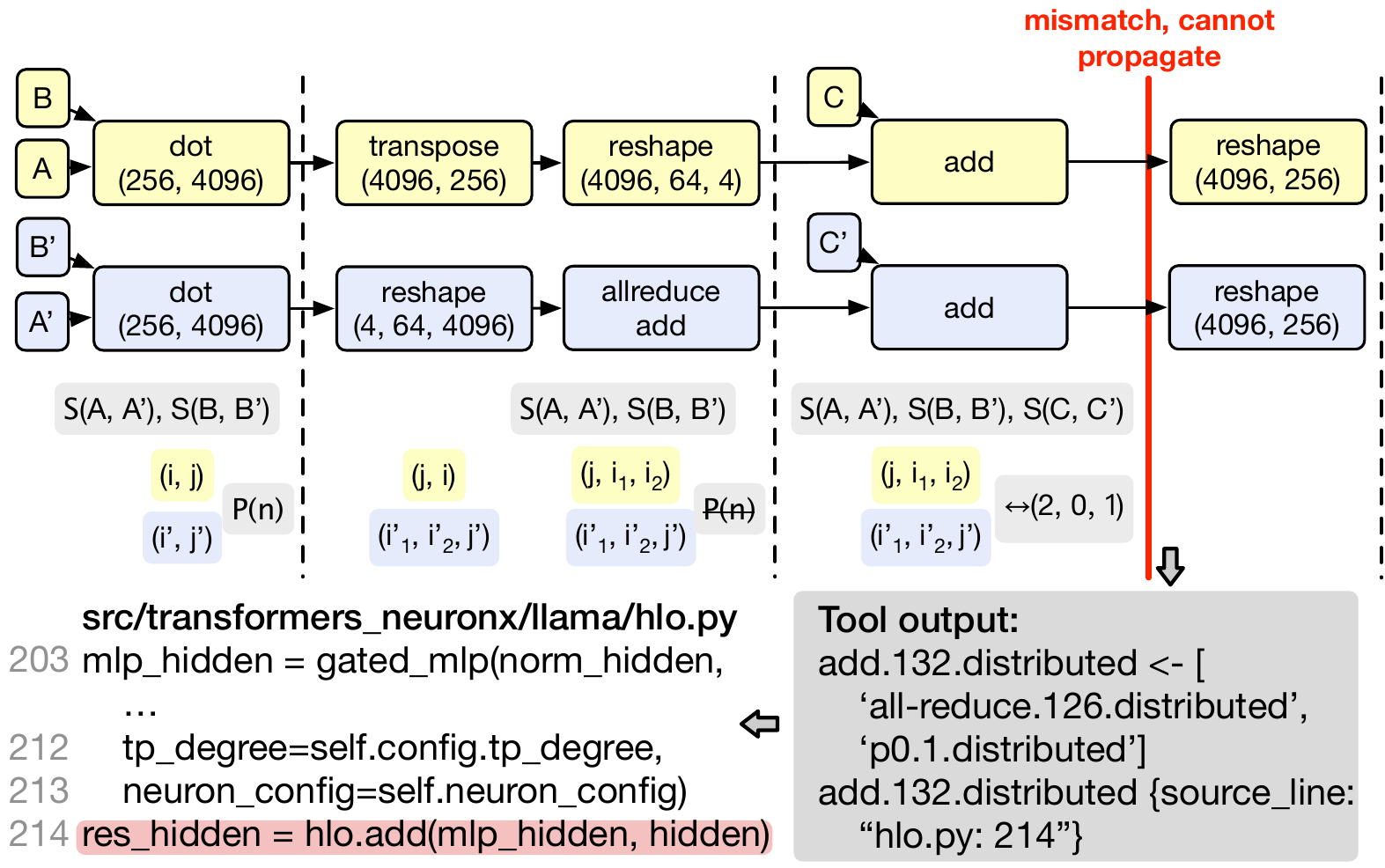}
   \caption{Upper graph is the oracle and bottom graph is the buggy graph. The layout relation does not propagate because the bijections of the inputs of add are not the same.    } 
  \label{fig:unverified_nodes}
\end{figure}

\textbf{Example.}
As shown in Figure~\ref{fig:unverified_nodes}, the buggy \texttt{add} operators are unverified because their input layout relations mismatch: the \texttt{reshape} and \texttt{allreduce add} align via \texttt{[transpose(2, 0, 1), reshape(4096, 64, 4)]}, while tensors $C$ and $C'$ align via a different transpose \texttt{(2, 1, 0)}. This mismatch prevents \sys from reasoning about the \texttt{add} nodes. \sys then reports these unverified \texttt{add} nodes and their consumers, highlighting verified inputs and revealing that the true issue originates from the \texttt{mlp\_hidden} variable after its \texttt{reshape}. Thus, the \texttt{add} node is pinpointed as the root of the discrepancy: its inputs are verified, but the node itself fails to align.

\section{Implementation}
\label{sec:impl}

We implement \sys primarily in Python with approximately 9k SLOC,
of which about 6.5K LOC are devoted to manually encoding 25 meta rules across different forms of parallelism (\emph{e.g.}, tensor, expert) and supporting the corresponding IR graph operators. We found the defined rules highly reusable across different semantic types.
Once the framework and the basic rule set are in place, adding rule support for new parallelism techniques requires only modest effort. For example, we added 2 rules to enable sequence parallelism, which are implemented in 30 LOC.

The core layout-equivalence reasoning is powered by egglog, which we integrate as the e-graph engine for tensor relation resolution.
\sys is built on top of PyTorch XLA to operate directly on the intermediate representation of ML models. Our prototype targets AWS Transformers NeuronX~\cite{transformersneuronx} and NeuronX Distributed~\cite{nxd} from the AWS Neuron SDK, though the core algorithms %
are framework-agnostic and could be ported to other IR-based systems (\emph{e.g.,} TensorFlow XLA, Megatron-LM) with moderate  effort.

\section{Evaluation}

We answer these questions in our evaluation:
(1) Does \sys support real-world large ML models (\S\ref{sec:eval:models})?
(2) Does \sys scale well as the size of models increases (\S\ref{sec:eval:scalabilty})?
(3) Does \sys detect any bugs in the real-world systems and what categories of bugs does \sys cover (\S\ref{sec:eval:bugs})?

\subsection{Verifying Large Real-world Models}
\label{sec:eval:models}
\textbf{Setup.}
The graphs are generated using the XLA compiler on AWS Trainium machines with 32 Neuron cores (Transformers NeuronX, version 0.12.313). Verification is performed on a 6-core AMD Ryzen 5 5600U CPU with 16 GB RAM.
Note that our evaluation is conducted on inference graphs generated by NeuronX. This choice is due to current framework support, not a limitation of \sys itself.

We evaluate our tool on two popular LLMs: Llama-3, representing dense models, and Mixtral, representing Mixture-of-Experts (MoE) architectures. We run the models with the parallelization pipelines commonly used by developers, including tensor parallelism. For Llama-3, we additionally verify equivalence under flash decoding and sequence parallelism, while for Mixtral we check expert parallelism implemented with recursive loops. In total, our evaluation covers four parallelism techniques. All generated graphs are executed on 32 Neuron cores, the largest configuration supported by NeuronX.

\begin{table}[h]
\caption{Evaluated real-world large models.}
\footnotesize
\centering
\begin{tabular}{@{}c c c c @{}}
\toprule
Exp. ID & Model  & Layers & Verification time      \\ \midrule
L1 & Llama-3.1-8B   & 32 & 48s  \\ 
L2 & Llama-3.1-70B  & 80 & 1m 40s  \\ 
L3 & Llama-3.1-405B & 126 & 2m 37s    \\ 
M1 & Mixtral-8x7B   & 32 & 1m 52s \\ 
M2 & Mixtral-8x22B  & 56 & 3m 1s \\ \bottomrule
\end{tabular}%
\label{tab:results}
\end{table}

\textbf{Results.}
We show the results in Table~\ref{tab:results}. All verification completes within a few minutes on a commodity machine, showing the practicality of our tool.
\sys takes longer time for Mixtral compared to Llama-3.1 since Mixtral models have more nodes and more fine-grained analysis (unroll loop of recursive add operations for expert parallelism). 
Unlike TrainVerify~\cite{trainverifySOSP2025}, which reasons at the per-element level across devices and takes days even for Llama-3.1-405B, \sys reasons over distributed tensors as sharded entities, yielding orders-of-magnitude faster runtimes.

\subsection{Scalability}
\label{sec:eval:scalabilty}

We evaluate the performance of \sys against 5 groups of parallelization and model configurations. The base model is the Llama-3.1-8B and the distributed graph is run across 32 neuron cores on AWS trainium machines. Table \ref{tab:scalability_config} represents the control variable and inference configurations for each group. The graphs are generated from Transformers NeuronX.

\begin{table}[h]
\caption{Configurations for scalability experiments.}
\footnotesize
\centering
\begin{tabular}{@{}c c c c c c c@{}}
\toprule
Group & Controlled Variable  & seqlen & bs & layers & tp & heads      \\ \midrule
a & sequence length  & * & 4 & 32 & 32 & 32  \\ 
b & batch size & 64 & * & 32 & 32 & 32  \\ 
c & layers & 64 & 4 & * & 32 & 32    \\ 
d & tensor parallelism & 64 & 4 & 32 & * & 32   \\ 
e & heads & 64 & 4 & 32 & 32 & * \\ \bottomrule

\end{tabular}%
\label{tab:scalability_config}
\end{table}

\begin{figure*}
\label{fig:scalability}
     \centering
              \includegraphics[width=\linewidth]{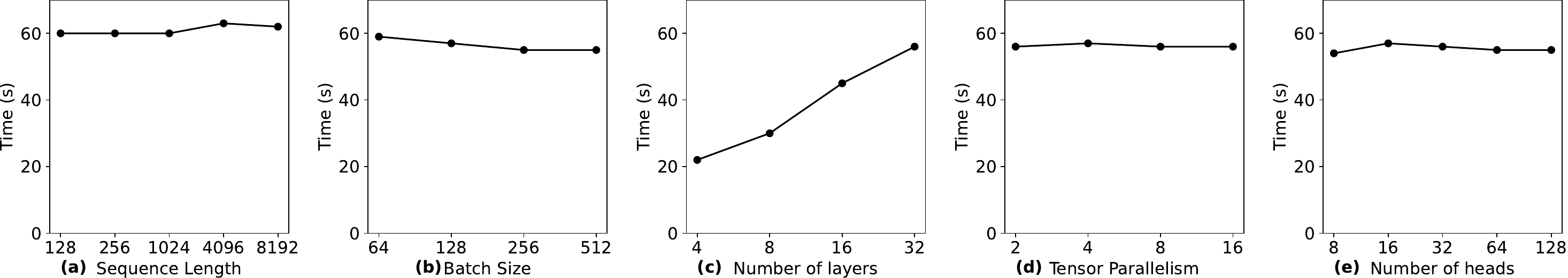}
     \caption{\sys's performance when checking the equivalence for different  configurations. }
    \label{fig:different_conf}
\end{figure*}

\begin{figure}
     \centering
              \includegraphics[width=0.8\linewidth]{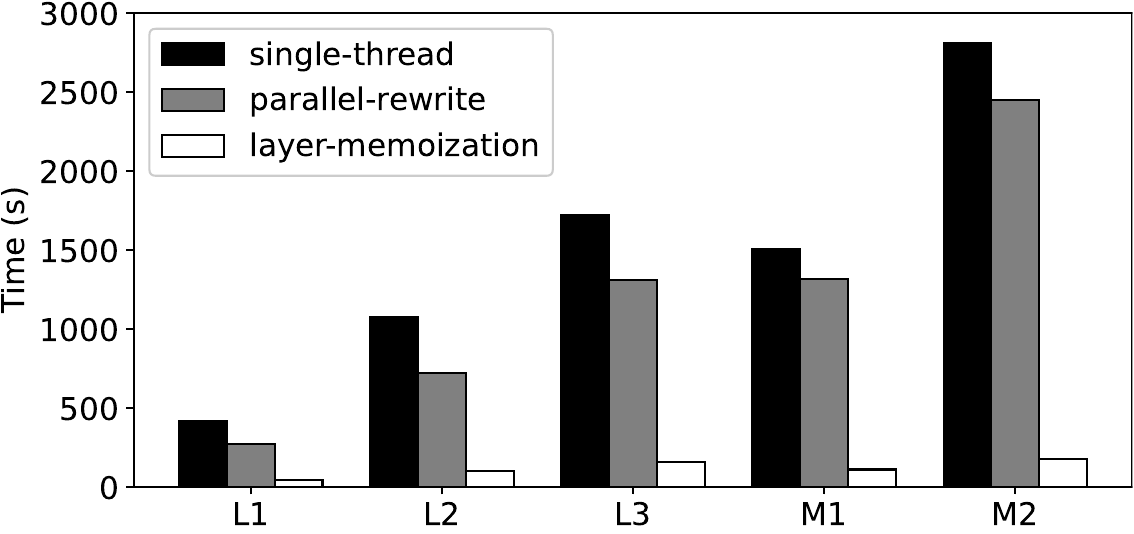}
   \caption{Verification time based on scaling techniques. }
   \label{fig:scalability}

\end{figure}

\textbf{Constant relationship to tensor shapes}. Shown in Figure \ref{fig:different_conf}a, \ref{fig:different_conf}b and \ref{fig:different_conf}e, the verification time is the same regardless of the size of the input tensor shapes. This is because \sys operates on the computation graph level, where the size of the expression stays the same regardless of the tensor shape. 

\textbf{Constant relationship to parallelization degree}. Shown in Figure \ref{fig:different_conf}d, the verification time of \sys is constant to the complexity of the parallelization since the size of the generated computation graph is the same regardless of the number of cores used to run the ML pipeline. The additional communication nodes introduced through these parallelization has a negligible effect to the total verification time. 

\textbf{Linear relationship to number of layers}. Shown in Figure \ref{fig:different_conf}c, increasing model layers causes the operators to be added sequentially, which causes the verification time of \sys to increase. The additional time is introduced due to the partitioning of graph into multiple subgraphs. 

\textbf{Graph partitioning and layer memoization verification}. We demonstrate the benefit of layer memoization by comparing it against sequential graph partitioning and parallel rewrite with graph partitioning as shown in Figure \ref{fig:scalability}. We run the experiments on Llama-3.1-8B model with TP degree of 32 and 32 layers.
We tried running the tool against the base Llama-3.1-8B model without graph partitioning and our tool fails to rewrite the entire model due to insufficient  resources.

\subsection{Bugs Detection}
\label{sec:eval:bugs}

First, we categorize classes of silent errors into the following categories that can be detected by \sys.

\begin{enumerate}[noitemsep, topsep=0pt, partopsep=0pt, leftmargin=*]
  \item \textbf{Incorrect distributed operation}: Using the wrong communication primitive (\emph{e.g.,}  an unnecessary all-reduce).
  \item \textbf{Incorrect distributed configuration}: Incorrect device assignment even when the operator itself is correct (\emph{e.g.,} reducing on only part of the cores).
  \item \textbf{Inconsistent tensor precision}: Single-device and distributed pipelines use different precisions.
  \item \textbf{Incorrect axis splitting}: Reshape operations split tensors incorrectly, breaking sharding relations.
  \item \textbf{Incorrect layout optimization}: An invalid sequence of layout transformations compared to the oracle.
\end{enumerate}

Categories 1-4 are covered by \sys's partition and layout analysis, which focuses on propagating relations through intermediate tensors whenever the tensors match the rules defined in Table \ref{tab:rule_templates}. 
Category 5 is eliminated through \sys's bijection inference. %

\begin{table}[h!]
\caption{Reproduced old bugs detected by \sys. \locpinpoint: pinpoint the faulty instr. \locfunc: pinpoint the faulty function or data structure. 
n/a: not applicable (undetected). 
}
\footnotesize
\centering
\begin{tabular}{@{}c l l l@{}}
\toprule
Bug ID & Bug description & Issue links & Loc. \\ \midrule
Bug\#1  & Incorrect layout optimization  & \cite{transformersneuronx_69d039d} & \locfunc \\
Bug\#2 & Incorrect all-to-all layout & \cite{deepspeed_5808}   & \locpinpoint \\
Bug\#3--6 & Missing all-reduce              & \cite{megatronlm_1699, megatronlm_5fffdfc, megatronlm_599, deepspeed_7188} & \locfunc \\
Bug\#7--8 & Missing normalization           & \cite{megatronlm_1620, megatronlm_1611} & \locfunc \\
Bug\#9--12 & Redundant all-reduce          & \cite{deepspeed_6714, nemo_8487, nemo_9344, transformerengine_3}   & \locpinpoint \\
Bug\#13--16 & Incorrect distributed replica groups & \cite{megatronlm_32bbb76, transformerengine_335, nemo_5564, deepspeed_5618}  & \locpinpoint \\
Bug\#17 & Inconsistent precision & \cite{deepspeed_2071} & \locpinpoint \\
Bug\#18 & Incorrect KV cache slicing & \cite{transformersneuronx_e2f5241} & n/a \\ 
Bug\#19 & Incorrect logits layout & \cite{transformersneuronx_0c646b0} & n/a \\ \bottomrule
\end{tabular}%
\label{tab:bugs_old}
\end{table}

\textbf{Setup.} 
We evaluate \sys on incorrect distributed computation graphs by reproducing 19 real-world bug cases across diverse ML systems, re-implemented in Transformers NeuronX.
These bugs represent real-world silent errors that are frequently encountered.
We use the Llama-3.1-8B model as workloads with TP degree of 32. 

\textbf{Finding reproduced old bugs.} 
The result is shown in Table~\ref{tab:bugs_old}.
17 out of 19 reproduced bugs are detected by \sys under one minute. %
Besides detection, we also show how the output pinpoints the bug in the source code. 
We already showed Bug\#1 in Figure \ref{fig:unverified_nodes} and explained in \S\ref{sec:design:mapping}.
For Bug\#3--8, \sys pinpoints the faulty function but not the missing operation required. For instance, in missing all-reduce bugs, 
\sys flags add operators which consumes partial and sharded tensors (%
\texttt{hlo.add(mlp\_hidden, hidden)}) because 
the input tensors have different relations.
Bug\#18--19 cannot be detected by \sys  
since the bug manifests outside of the graph compilation phase.

\begin{table}[h!]
\caption{New bugs exposed by \sys. \locpinpoint: pinpoint the faulty instr. \locfunc: pinpoint the faulty function or data structure. 
}
\footnotesize
\centering
\begin{tabular}{@{}c l l l@{}}
\toprule
Bug ID & Bug description & Framework  & Loc. \\ \midrule
Bug\#1 & Incorrect layout optimization  & TNx~\cite{transformersneuronx}    & \locfunc \\
Bug\#2 & Wrong all-to-all transformation & TNx~\cite{transformersneuronx}        & \locpinpoint \\
Bug\#3 & Wrong sharding of tensors       & TNx~\cite{transformersneuronx}        & \locpinpoint \\
Bug\#4 & Wrong precision ordering        & NxD~\cite{nxd}        & \locfunc \\
Bug\#5 & Wrong operation ordering        & NxD~\cite{nxd}        & \locfunc \\ \bottomrule
\end{tabular}%
\label{tab:bugs_new}
\end{table}

\textbf{Exposing new bugs.} 
During our evaluation, \sys successfully uncovered five previously unknown bugs in the Amazon SDK.
All of them cause severe incorrectness issues.
For Bug \#1, \sys localized the fault to layout optimization rather than the intermediate reshape–transpose sequence, which it treats as verified states. For Bugs \#2–3, \sys pinpointed incorrect reshape operators in all-to-all transformations and tensor sharding, where reshapes failed to produce semantically equivalent tensors. For Bugs \#4–5, \sys identified the faulty functions but not the root cause; for example, Bug \#4 was localized to positional rotary embedding calculations, where inconsistent precision stemmed from the compiler rather than the PyTorch function itself.
We have submitted these bugs as well as reproducing scripts to developers.

\section{Limitations and Future Work}
\label{sec:limitation}

First, \sys focuses on the verification of computational graphs.
Bugs that do not manifest at graph IR level (scheduling bugs (\emph{e.g.,} data races in distributed pipelines) or runtime errors that occur before graph compilation) may still occur even when the computational graph is correct. 
Second, even only considering verification of computational graphs, \sys is sound (\emph{i.e.,} it guarantees the correctness of computational graphs that can be verified), but not complete (\emph{i.e.}, it offers no guarantee on verifying all correct computational graphs). 
Third, \sys supports popular parallelism techniques such as tensor parallelism and flash decoding, meanwhile it requires additional effort to extend support to more finer-grained  patterns (\emph{e.g.,} context/pipeline parallelism), which has more complex inter-device communication and runtime semantics. 
Additionally, we demonstrate \sys on inference graphs because our current toolchain exports inference IRs. The techniques themselves are agnostic to training vs. inference.
Finally, \sys pinpoints the lines where discrepancies arise but may not reveal the root cause, leaving manual diagnosis to developers---a step we aim to assist with LLMs.
\section{Related Work}
\label{sec:related}

\textbf{ML workflow reliability.}
Many works~\cite{bamboo2023nsdi,varuna22eurosys,checkfreq2021fast,gemini2023sosp,oobleck2023sosp} 
propose runtime recovery strategies to ensure ML workflows reliable. 
However, most of them focus on visible failures like crashes or hangs
and are vulnerable to subtle errors. 
This work focuses on exposing silent errors
inspired by related study in distributed systems~\cite{zodiac2024sosp,oathkeeper2022osdi,silent2023sosp,
gunawi16,GrayFailureHotOS2017, T2COSDI2025}.

\textbf{Equivalence checking.}
Checking semantic equivalence between programs~\cite{vericert2021oopsla,hls2024fpga,eqcheck2012toplas,ruler2021oopsla,enumo2023oopsla} has long been an important research topic. Equality saturation~\cite{es2009popl,egglog2023pldi}, widely used for program optimization, has recently been extended to machine learning~\cite{qgym2022pact,tree2024pact,superoptimization2021mlsys}.
PolyJuice~\cite{polyjuice2024oopsla} generates 
equivalent but syntactically different graphs with equality saturation 
to find mis-compilation bugs. 
We check equivalence of two existing graphs, and
 face distinct challenges.

Recent works~\cite{eleftheriadis2022neuralnetworkequivalencechecking, trainverifySOSP2025} propose to verify ML computation graphs with SMT solvers.
TrainVerify~\cite{trainverifySOSP2025} targets the same goal of verifying  ML graph equivalence but discharges proofs to SMT, ultimately relying on costly reasoning plus operator-specific reductions. It further assumes tensor lineage between logical and parallel graphs and offers limited localization when proofs fail. Our e-graph approach performs scalable equivalence checking and yields precise bug localization.

\textbf{Graph rewriting.}
Optimizing ML performance through computational graph 
rewriting~\cite{pet2021osdi,superoptimization2021mlsys,mirage2024arxiv,mcts-geb2023euromlsys} 
has gained significant attention in recent years. TASO~\cite{taso2019sosp} automates 
graph substitutions by leveraging formally verified operator specifications. 
Alpa~\cite{alpa2022osdi} optimizes computation graphs by automatically generating 
parallel execution plans at both the inter- and intra-operator levels. 
While their primary focus is 
performance optimization, these works ensure correctness by enforcing equivalence between 
transformations.

\textbf{ML framework testing.}
It is known that machine learning frameworks are error-prone to bugs due to their complexity.
Researchers  conduct bug studies~\cite{optimizebugstudy2023icse,compilerstudy2021fse,
tambon2023silentbugsdeeplearning} and propose testing techniques for 
deep learning libraries~\cite{eagle2022icse, deepxplore2017sosp, deeptest2018icse, TrainCheckOSDI2025}
and compilers~\cite{tzer2022oopsla,polyjuice2024oopsla, mlirsmith2023ase}.
NNSmith~\cite{nnsmith2023asplos}
generates diverse DNN test models and uses differential testing to identify bugs.
Meanwhile, testing cannot eliminate all bugs and issues still escape from checking.
\sys aims to capture those missing issues before they manifest at runtime.

\textbf{Verified compilers.}
Compiler verification has been a promising direction to ensure
the correctness of compiled tensor programs.
There have been efforts verifying the soundness of 
rewrite rules in ATL~\cite{verified-atl2024pldi}, XLA~\cite{tensorright2025popl}
and Halide compiler~\cite{verified-halide2022oopsla}.
Our approach is complementary to their efforts by verifying 
at generated IR graph level. %

\section{Conclusion}
\label{sec:conclusion}

We presented \sys, a lightweight verifier that checks semantic equivalence between baseline and distributed IR graphs using e-graph rewriting, relation propagation, and symbolic bijection inference. On Llama-3.1 and Mixtral variants, \sys validates full graphs in minutes on a commodity machine, detects 17 out of 19 historical bugs, and uncovers 5 new ones. Our results suggest  computational graph verification is practical for large machine learning models. %

{
\bibliographystyle{ACM-Reference-Format}
\bibliography{bib/reference,bib/study,bib/gray,bib/cloud,bib/incidents,bib/bugs,bib/detection,bib/leak,bib/ml,bib/verification}
}

\end{document}